# ASR Context-Sensitive Error Correction Based on Microsoft N-Gram Dataset

Youssef Bassil, Paul Semaan

**Abstract**—At the present time, computers are employed to solve complex tasks and problems ranging from simple calculations to intensive digital image processing and intricate algorithmic optimization problems to computationally-demanding weather forecasting problems. ASR short for Automatic Speech Recognition is yet another type of computational problem whose purpose is to recognize human spoken speech and convert it into text that can be processed by a computer. Despite that ASR has many versatile and pervasive real-world applications,it is still relatively erroneous and not perfectly solved as it is prone to produce spelling errors in the recognized text, especially if the ASR system is operating in a noisy environment, its vocabulary size is limited, and its input speech is of bad or low quality. This paper proposes a post-editing ASR error correction method based on MicrosoftN-Gram dataset for detecting and correcting spelling errors generated by ASR systems. The proposed method comprises an error detection algorithm for detecting word errors; a candidate corrections generation algorithm for generating correction suggestions for the detected word errors; and a context-sensitive error correction algorithm for selecting the best candidate for correction. The virtue of using the Microsoft N-Gram dataset is that it contains real-world data and word sequences extracted from the web which canmimica comprehensive dictionary of words having a large and all-inclusive vocabulary. Experiments conducted on numerous speeches, performed by different speakers, showed a remarkable reduction in ASR errors. Future research can improve upon the proposed algorithm so much so that it can be parallelized to take advantage of multiprocessor and distributed systems.

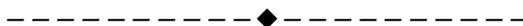

## 1 INTRODUCTION

With the advancement of information technologies, computers are no more exclusively used for performing mathematical and scientific operations. Instead, miscellaneous applications are now possible, allowing computers to solve versatile problems pertaining to different fields and domains. ASR short for Automatic Speech Recognition has been a subject of great focus and attention in recent years as it has been studied and researched by severalscientists, universities, and research centers. Inherently, ASR converts spoken words represented mathematically as an acoustic waveform into text that can be processed by a computer [1].Speech-To-Text (STT), Automated Telephone Services (ATS), Voice User Interface (VUI), Voice-driven Home Automation (Domotics), and Speech Dictation are few ASR applications to mention.

In spite of the great advantages and benefits of ASR, it isstill error-prone and imperfect as it produces spelling errors in the recognized output text. Commonly, ASR errors are manifested as linguistic mistakes and misspellings visible at the final output of the system. These errors are often caused by the extreme noise in theenvironment, the bad quality of the speech, the fluctuating utterance of the dialogue, and the small size of

- *Youssef Bassil is the Chief Science Officer of the Lebanese Association for Computational Sciences, (LACSC), reg. no. 957, 2011, Beirut, Lebanon.*

- *Paul Semaan is a senior researcher at the Lebanese Association for Computational Sciences, (LACSC), reg. no. 957, 2011, Beirut, Lebanon.*

the ASR vocabulary [2], [3].

Numerous error-correction methods and algorithms were devised to help fight against ASR errors, some of themrely on post-processing the output text and correcting it manually;whereas others rely on building improved acoustic models to increase the precision of speech recognition [4]. Regardless of all these attempts focused on reducing the ASR error rate, results are not yet convincing and speech recognition systems still suffer a major degradation in performance.

This paper proposes a post-editing ASR error correction method for detecting and correcting non-word and real-word errors generated by ASR recognition systems, based on data extracted from Microsoft Web N-Gram dataset [5]. Principally, the Microsoft Web N-Gram datasetenclosespetabytes ofn-gram word counts and statistics retrieved from the Internet and Bing search engine [6], and is appropriate for carrying out text spelling correction. The proposed approach is a post-editing process which spell-checks the final recognized output text after the input wave has been completely converted.Itis majorly composed of three foremost algorithms: An error detection algorithmthat detects non-word errors in the ASRoutput text using unigram information from Microsoft Web N-Gram dataset; a candidate corrections generationalgorithmthatgenerates possible correction spellings for the misspelled words using a character-based2-gram model; and a context-sensitivereal-word error correction algorithmthatpicks out the closest candidate for correction using 5-gram counts from Microsoft Web N-Gram dataset. In effect, as the proposed techniquemakes use of real-world web-scale data, it can significantly decrease the ASR error rate and consequently improve theperformance of ASR systems.

## 2 AUTOMATIC SPEECH RECOGNITION

As defined by many textbooks [7], [8], and [9], automatic speech recognition systems also known as ASR, receive some speech signals as input and generate a corresponding readable text transcript as output. Put differently, it simply converts spoken words into text. Figure 1 shows a speech recognition system in which a voice W usually generated by a speaker such a human person, propagates as a waveform into the communication channel where it is analyzed and processed to eventually be transformed into a readable text W'.

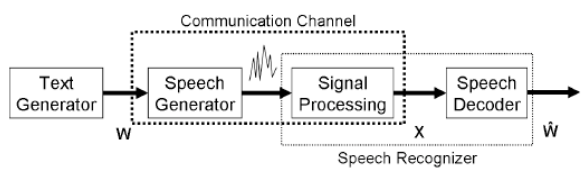

Fig. 1. A Basic Speech Recognition System

Essentially, an ASR system is often implemented using a Hidden Markov Model (HMM) [10], [11] based on the notion of noisy channel [12]. The concept behind the noisy channel model is to consider the input acoustic waveform as a noisy signal which has been distorted somehow during transmission. The quintessence of this approach is that if one could know how the original waveform was distorted, it is then easy to find the original input.The Hidden Markov Model is a special type of weighted finite-state automata, defined by a set of $N$ states $Q=q_1 q_1 \ldots q_n$; a start state $q_0$ and an end state $q_f$; a sequence of input observations $O=o_1 o_2 \ldots o_t$; a set of transitions from one state to another based on the input observations; and two types of probabilities: The prior probability and the likelihood probability. The former is associated with every transition and indicates how likely a transition is to be taken.It is represented by a transition probability matrix $A=a_{11} a_{12} \ldots a_{n1} \ldots a_{nm}$ where $a_{ij}$ represents the probability of transiting from state $i$ to state $j$. The latter is denoted by $B=b_i(o_t)$, and consists of a sequence of observations likelihood that indicates the probability of an observation $o_i$ being emerged from state $i$.

Fundamentally, an ASR system is a blend of four logical modules [7], each of which has a particular algorithm, purpose, and inner-workings, and they are in order: The signal processing module, the acoustic modeling module, the language modeling module, and the decoding module. Figure 2 depicts a block diagram of an automatic speech recognition system comprising four functional modules.

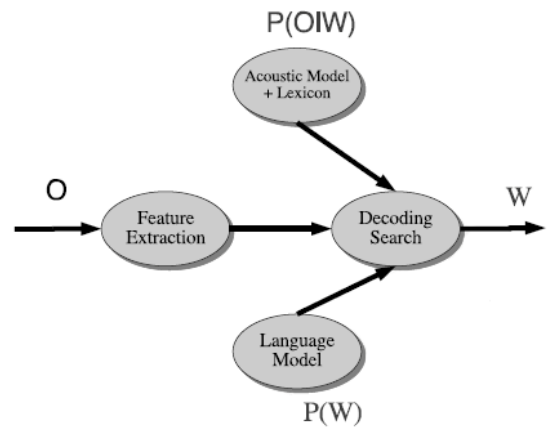

Fig. 2. Block Diagram of an ASR System

The first is the signal processing module, in which spectral features are extracted from the input acoustic waveform by sampling and capturing small frames out of the input signal on a maximum interval of 20 milliseconds. The spectral features help building phone and sub-phone classifiers. Phones are individual symbols that model the pronunciation of a word.Computationally, the input signal is first converted from analog to digital, then the power of its high frequencies is amplified in a process called Pre-emphasis with the objective of increasing the speech detection accuracy. Afterwards, the windowing process is introduced to divide the signal into frames of signal speeches having a particular length usually 25ms. These frames are separated from each other by an offset called frame shift usually of length 10ms. Next, the Discrete Fourier Transform (DFT) is applied to transform the previous signal frames into a complex number representing the phase and magnitude of the frequency of that frame. Finally, the cepstrum is calculated via the Inverse DFT which significantly improves the accuracy and performance of phones recognition. The result of this module is 39 features called Mel Frequency Cepstral Coefficients (MFCC), which uniquely identify a discrete acoustic phone in the input sound.

The second is the acoustic modeling (AM) module, which computes the likelihood of the observed spectral feature vectors given linguistic units (phones).For instance, it computes the likelihood $P(o|q)$ of a specific feature vector $o$ given a particular HMM stage $q$ that represents a particular phone $x$. For this purpose, a lexicon of words with their corresponding phones (a sequence of pronunciations), is used to help recognizing the spoken words.

The third is the language modeling (LM) module, which computes the prior probability $P(W)$that approximates how likely a given sentence is to occur in the language. $P(W)$ is usually calculated using the probabilistic $n$-gram model [13] which predicts the next word, letter, or phone in a given sequence. In short, an $n$-gram is simply a collocation of words that is $n$ words long.

The fourth is the decoding module, which joins the observation likelihood $P(O|W)$ resulted from of the acoustic model and the prior probability $P(W)$ resulted from the language model to deduce the most likely output text $W'$. The products of all probabilities are calculated $P(O|W)*P(W)$ and the one with the greatest value is selected as the output text $W'$. Figure 3 is the mathematical equation for calculating and choosing the most probable output text $W'$.

$$\dot{W} = \underset{W \in \mathscr{L}}{\mathrm{argmax}} \overbrace{P(O|W)}^{\text{likelihood}} \overbrace{P(W)}^{\text{prior}}$$

Fig.3.Equation for Finding the Product of the Likelihood and the Prior Probability

## 3 ERRORS IN SPEECH RECOGNITION SYSTEMS

An early experiment conducted at IBM research labs [14] to calculate the number of errors generated by ASR systems, showed an average of 105 errors being committed per minute.Essentially, ASR errors are relatively due to five factors:The first factor is noise which is tightly associated withthe condition of the location in which speech recognition is being performed.For instance, if the recognition process occurs in a silent environment, it yieldsto more accurate results than if it would have been occurred in a noisy environment. Noise adds extra signals to the speech; and hence,it alters the content of the input waveform making it hard to be interpreted.The second factor is the type of speech being recognized. In fact, there exist two types of speech: thediscrete speechalso called isolated-word speech in which spoken words are separated by silent pauses, andthe continuous speech which contains non-segmented endless sequence of words that are more difficult to be separated and distinguished by the system. Continuous speech thus imposes more complications on the recognitionprocess whichresult in an increase in the *ASR*error rate. The third factor is the speech utterance which ranges from read speech to conversational speech. Conversational speech is more problematic to handle since it is spontaneous and may contain defects in pronunciation.The fourth factor is the dialect of thespeech which varies from speaker to speaker as each person has unique spectral features. ASR systems can sometimes be characterized as speaker-dependentand speaker-independent systems. A speaker-independent system operates for different speakers and for different type of speakers. Such systems are more challenging to develop and implement, and are subject to higher error rate.The fifth factor is the size of the vocabulary that anASR system can recognize. Basically, and since the acoustic model (AM) is based on an internal dictionaryor lexicon of words with their corresponding phones and pronunciations that are necessary to match a spoken word with an entry in the lexicon, the larger the size of the vocabulary thelexiconhas, the less is to be the ASR error rate. A small vocabulary can lead to a situation often known as OOV short for Out-Of-Vocabulary which usually occurs when a spoken word is not found in the acoustic model's dictionary.This would in consequence cause theASR system to failrecognizingthe input vocal word.

## 4 LIMITATIONS OF ASR BUILT-INDICTIONARIES

The vocabulary size and the number of distinct words that an ASRsystem can recognize play a turning point in determining the overall error rate of the system. Applications with fewer terms like "yes" and "no", or digits like "1,2,3...9" are easier to handle than those with large volume of terms such as continuous dictation systems that sometimes require the recognition of a hive of terms and words. Such systems are called LVCSR (Large-Vocabulary Continuous Speech Recognition) and they usually need to recognize between 20,000 and 60,000 terms while achieving a good level of accuracy and a minimal amount of errors. Since ASR systemsare often built on traditional dictionariesthat do not cover all wordsin the language, theysuffer from severe Out-Of-Vocabulary(OOV) deficiencies. The reasons behind OOV can be summarized by the following:

The first reason is thatASRsystems lack a comprehensive dictionary that can cover every single word in the language. For instance, the Oxford dictionary embraces 171,476 words in current use, and 47,156 obsolete words, in addition to their derivatives which count around 9,500 words. This suggests that there is, at the very least, a quarter of a million distinct English words. Besides, spoken languages may have one or more varieties each with dissimilar words, for instance, the German language has two varieties, a new-spelling variance and an old-spelling variance. Likewise, the Armenian language has three varieties each with a number of deviating words: Eastern Armenian, Western Armenian, and Grabar. Therefore, it is obvious that languages are not uniform, in a sense that they are not standardized and thereby cannot be supported by a single dictionary.

The second reason is that regular dictionaries normally target a specific language in that they cannot support multiple languages simultaneously. For instance, the Oxford dictionary only targets the English language. The Hachette dictionary targets the French language, while the Al Kamel dictionary targets the Arabic language. Therefore, it is unquestionably impossible to create an international dictionary pertaining to all languages of the world.

The third reason is that conventional dictionaries do not expansively support proper and personal names, names of countries, regions, geographical locations, technical keywords, domain specific terms, and acronyms. For instance, an ordinary dictionary can falsely detect "Andrew Jackson", "Intel", and "Renault" as incorrect words. Relatedly, technical terminologies such as "USB", "SATA", and "Texel", and names of diseases such as "Leukemia", "Parkinson", and "Cholera" canbe falsely detected as misspellings too. In total, it is nearly impracticable to compile a universal lexicon containing words from all existing domains and fields.

The fourth and last reason is that the content of standard dictionaries is static in a way that it is not constantly updated with new emerging words unless manually edited, and thus, it cannot keep pace with the immense dynamic breeding of new words and terms.

For all the aforementioned reasons, attaining better speech recognition results greatly require finding a universal, all-inclusive, multi-language, and dynamic dictionary embracing a colossal volume of real-world entries, words, terms, proper nouns, expressions, jargons, and terminologies.

## 5 State-of-the-Art ASR Error Correction Techniques

Different error correction techniques exist, whose purpose is to detect and correct misspelled words generated by ASR systems. Broadly, they can be broken down into several categories: Manual error correction, error correction based on alternative hypothesis, error correction based on pattern learning, and post-editing error correction.

In manual error correction, a staff of people is hired to review the output transcript generated by the ASR system and correct the misspelled words manually by hand. This is to some extent considered laborious, time consuming, and error-prone as the human eye may miss some errors.

Another category of error correction is the alternative hypothesis in which an error is replaced by an alternative word-correction called hypothesis. The chief drawback of this method is that the hypothesis is usually derived from a lexicon of words; and hence, it is susceptible to high out-of-vocabulary rate. In that context, Setlur, Sukkar, andJacob[15] proposed an algorithm that treats each utterance of the spoken word as hypothesis and assigns it a confidence score during the recognition. The hypothesis which bypasses a specific threshold is to be selected as the correct output word. The experiments showed that the error rate was reduced by a factor of 0.13%. Likewise, Zhou, Meng, and Lo[16] proposed another algorithm to detect and correct misspellings in ASR systems. In this approach, twenty alternative words are generated for every single word error and treated as utterance hypotheses. Then, a linear scoring system is used to score every utterance with certain mutual information representing the frequency or the number of occurrence of this specific utterance in the input waveform. Next, utterances are ranked according to their scores. The utterance that received the highest score is chosen to substitute the detected error. Experiments conducted, indicated a decrease in the error rate by a factor of 0.8%.

Pattern learning error correction is yet another type of error correction techniques in which error detection is done through finding patterns that are considered erroneous. The system is first trained using a set of error words belonging to a specific domain. Subsequently, the system builds up detection rules that can pinpoint errors once they occur. At recognition time, the ASR system can detect linguistic errors by validating the output text against its predefined rules. The drawback of this approach is that it is domain specific; and thus, the number of words that can be recognized by the system is minimal. In this perspective, Mangu and Padmanabhan[17] proposed a transformation-based learning algorithm for ASR error correction. The algorithm exploits confusion network to learn error patterns while the system is offline. At run-time, these learned rules assist in selecting an alternative correction to replace the detected error. Similarly, Kaki, Sumita, and H. Iida[18] proposed an error correction algorithm based on pattern learning to detect misspellings and on similarity string matching algorithm to correct misspellings. In this technique, the output recognized transcript is searched for potential misspelled words. Once an error pattern is detected, the similarity string algorithm is applied to suggest a correction for the error word. Experiments were executed on a Japanese speech and the results indicated an overall 8.5% reduction in ASR errors. In a parallel effort, statistical-based pattern learning techniques were also developed.Jung, Jeong, and Lee[19] employed the noisy channel model to detect error patterns in the output text. Unlike other pattern learning techniques which exploit word tokens, this approach applies pattern learning on smaller units, namely individual characters. The global outcome was a 40% improvement in the error correction rate. Furthermore, Sarma and Palmer[20] proposed a method for detecting errors based on statistical co-occurrence of words in the output transcript. The idea revolves around contextual information which states that a word usually appears in a text with some highly co-occurred words. As a result, if an error occurs within a specific set of words, the correction can be statistically deduced from the co-occurred words that often appear in the same set.

The final type of error correction is post-editing. In this approach, an extra layer is appended to the ASR system with the intention of detecting and correcting misspellings in the final output text after the recognition of the speech is completed. The advantage of this technique is that it is loosely coupled with the inner signal and recognition algorithms of the ASR system; and thus, it is easy to be implemented and integrated into an existing ASR system and can also benefit from other error correction explorations done in sister fields such as OCR, NLP, and machine translation. As an initial attempt, Ringger and Allen[21] proposed a post-processor model for discovering statistical error patterns and correct errors. The post-processor was trained on data from a specific domain to spell-check articles belonging to the same domain. The actual design is composed of a channel model to detect errors generated during the speech recognition phase, and a language model to provide spelling suggestions for those detected errors. As outcome, around 20% improvement in the error correction rate was achieved. On the other hand, Ringger and Allen[22] proposed a post-editing model named SPEECHPP to correct word errors generated by ASR systems. The model uses a noisy channel to detect and correct errors; it also uses the Viterbi search algorithm to implement the language model. Moreover, the system leverages a fertility model to deal with split or merged errors such as 1-to-2 or 2-to-1 mapping errors. Another attempt was presented by Brandow and Strzalkowski[23] in whichthe text generated by the ASR system is collected and aligned with the correct transcription of the same text. In a training process, a set of correction rules are generated from these transcription texts and validated against a generic corpus.The Rules that are void or invalid are discarded. The system loops for several iterations until all rules get verified. Finally, a post-editing stage is employed which exploits these rules to detect and correct misspelled words generated by theASRsystem.

## 6 PROPOSED METHOD

This paper proposes a novelpost-editing ASRcontext-sensitive error correction method based on Microsoft Web N-Gram dataset [5] for detecting and correcting non-word and real-word errors produced byASRsystems. The proposed method uses a post-editing approach in that it spell-checks the output transcript of the ASR system after the input speech has been converted into text. Microsoft who owns Bing search engine [6] already developed and published a set of online public APIs and Web Services to give access to their indexed web data. The Microsoft Web N-Gram dataset is a database containingreal-world web-scale dynamic data stored as word $n$-gram sequences with their corresponding counts,worthwhile in solving manycomputational linguisticsproblems. Since Microsoft dataset houses a huge volume of data crawled from real-world web pages and documents postedon the Internet, it is overflowedwith proper names, technical keywords, domain specific terms, acronyms, special expressions, and terminologies,that altogether can mimic a wide-ranging dictionary thatcan covermost of the words in the language.

Predominantly, the proposed error correction method combines three algorithms: The error detection algorithm,the candidate corrections generationalgorithm, and thecontext-sensitive error correction algorithm. Figure 4shows thelogical block diagramfor the proposed ASR error correction method.

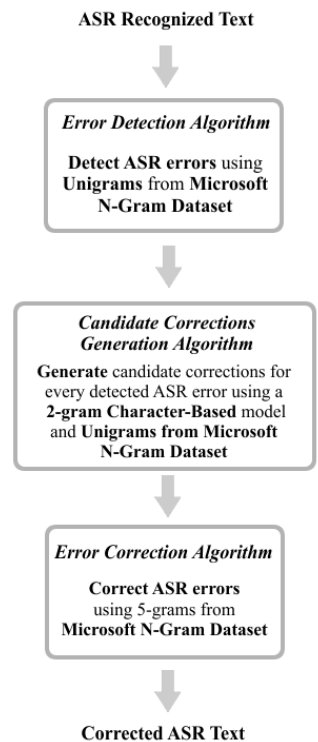

<div align="center">

ASR Recognized Text

**Error Detection Algorithm**

**Detect ASR errors using Unigrams** from **Microsoft N-Gram Dataset**

**Candidate Corrections Generation Algorithm**

**Generate** candidate corrections for every detected ASR error using a **2-gram Character-Based** model and **Unigrams from Microsoft N-Gram Dataset**

**Error Correction Algorithm**

**Correct ASR errors using 5-grams from Microsoft N-Gram Dataset**

Corrected ASR Text

</div>

Fig. 4.Proposed ASR Error Correction Method

### 6.1 The Error Detection Algorithm

The error detectionalgorithmdetects non-word spelling errors in the ASR output text. Formally, these errors are denoted by$E=\{e_1,e_2,e_3,e_p\}$ where $e$denotes an non-word error, and $p$ denotes the total number of detected errors.The ASR output text is denoted by$A=\{a_1,a_2,a_3,a_t\}$ where$a$ is a word or term in the ASRoutput text, and $t$is the total number of words. The algorithm works as follows: it first starts by validating every word $a_i$ in $A$against Microsoft Web N-Gram dataset; if an entry for $a_i$ is found in the dataset, then $a_i$ is assumedto bespelled correctly, and therefore no spelling correction is required. In contrast, if no entry exists for the word $a_i$in the dataset, then $a_i$ is assumed to be misspelled, and thusit requires a spelling correction. In due course,all the detectedspelling errors are grouped in a list,denoted by$E=\{e_1,e_2,e_3,e_p\}$ where $p$ is the total number of non-word errors detected in the $ASR$output text. The pseudo-codefor the error detection algorithm is given below.

**Function**ErrorDetection (A)
{
*// split the ASR text on space and return word tokens*
W←Split(A,"")

**for**(i←0 to i < N)  *// detect all word tokens*
    {
        *// search for W[i] in Microsoft N-Gram dataset*
R←Search(MicrosoftDataset , W[i])

**if**(R == true)  *// mean W[i] was found in Microsoft dataset(i.e. correctly spelled)*

i← i+1  // go to the next word tokenW[i+1]

**else**  // W[i] is misspelled and thus a correction is required
// go to the candidate corrections generation algorithm
GenerateCandidates(W[i])
}
}

## 6.2 The Candidate Corrections Generation Algorithm

The candidate corrections generationalgorithm generates a list of possible spelling correctionsfor the errors that were previously detected by the error detection algorithm. The list of candidates is denoted by $C=\{c_{11},c_{12},c_{13},c_{1q},...,c_{b1},c_{b2},c_{b3},c_{bq}\}$ where $p$ denotes a particular candidate spelling, $b$ denotes the total number of detected non-word ASR errors, and $q$ denotes the total number of candidate corrections generated for a particular error. Computationally, the algorithm generates candidate corrections using a 2-gram character-based model thatsearches for unigrams in Microsoft Web N-Gram dataset having 2-gram character sequences in common with theerror word.

For example, considering a speech that was converted by theASR system into "watch episodes of your favorite shaws and more", in which the word "shows" was misrecognized as "shaws". Converting the word "shaws" into 2-gram character sequences would produce: "sh" , "ha" , "aw" , "ws". The task of the candidate corrections generation algorithmis to find a series of unigrams from Microsoft Web N-Gram datasetthat enclose these 2-gram sequences. Table 1 shows a possible set of unigrams for the error word "shaws" retrieved from Microsoft Web N-Gram dataset.

TABLE 1
UNIGRAMS SHARING 2-GRAM CHARACTERS WITH THE ERROR "SHAWS"

| 2-Gram Sequences | Unigrams from Microsoft Web N-Gram Data Set |
|---|---|
| sh | _sh_ows _sh_awls _sh_ays _sh_ank _sh_am |
| ha | s_ha_wls s_ha_ys s_ha_nk _ha_ws _ha_wk |
| aw | s_aw_s s_aw_n m_aw_s h_aw_s h_aw_k |
| ws | sho_ws_ sa_ws_ ma_ws_ ha_ws_ he_ws_ |

Now the task is to find the unigrams having the highest number of common 2-gram character sequences with the error word "shaws". As there might be hundreds of unigrams, the algorithm only selects the top 8 unigrams as candidate corrections. Table 2 outlines the list of the top 8 candidate corrections for the error word "shaws".

TABLE 2
LIST OF TOP 8 CANDIDATE CORRECTIONS

| Candidate Corrections | Number of Common 2-Gram Character Sequences |
|---|---|
| haws | 3 |
| saws | 2 |
| hawk | 2 |
| shows | 2 |
| shawls | 2 |
| shays | 2 |
| shank | 2 |
| maws | 2 |

The pseudo-code of the candidate corrections generation algorithm is given below.

**Function**GenerateCandidates (word)
{
// create 2-gram character sequences and put them in a
a←Split2Grams(word)

**for**(i←0 to i < N)  // for all 2-gram sequences
   {
// look for unigrams having a[i] as substring, (i.e.
//unigrams sharing 2-gram sequence with the error word

L[i] ←Substring(MicrosoftDataset, a[i])

i← i+1
}

   // select the top 8 unigrams sharing 2-gram character
sequences with the error word
candidates←commonUnigrams(L)

// go to the error correction algorithm
ErrorCorrection(candidates)
}

## 6.3 The Error Correction Algorithm

The error correctionalgorithm selects the best candidate spelling as a correction for every detected error. The algorithm first starts by considering each generated candidate correction$c_{ir}$ with 4 words that directly precede the initial error in the ASR output text. The result is a 5 words sentence denoted as $L_r="a_{i-4}a_{i-3}a_{i-2}a_{i-1} \quad c_{ir}$ " where $L$denotes a sentence made out of 5 words, $a$ denotes a word preceding the original ASR error, $c$ denotes a particular candidate correction for a particular error, $i$ denotes the $i_{th}$ word that precedes the originalASR error, and $r$ denotes the $r_{th}$ candidate correction. Subsequently, the algorithm searches for every $L_r$ inMicrosoft Web N-Gram dataset. The candidate $c_{ir}$ in sentence $L_r$ having the highest number of occurrence in Microsoft Web N-Gram dataset is selected to replace the originally detected ASR error.

The proposed algorithm is context-sensitive as it depends on real-world data statistics from Microsoft Web N-Gram dataset, largelydug up from the Internet.

Accordingly, and back to the previous example, in spiteof the fact that the candidate "haws" is a valid correction for the error word "shaws", candidate "shows" will be favoredand selected as a spelling correction instead of "haws" since the sentence "watch episodes of your favorite <u>haws</u> and more" would occur fewer times over the Internet than the sentence "watch episodes of your favorite <u>shows</u> and more". Table 3 outlinesthe various$L_r$ 5-gram sentences from Microsoft Web N-Gram dataset, each enclosing a word from the list of candidate corrections.

TABLE 3
5-Gram Sentences from Microsoft Web N-Gram dataset

| r | $L_r$  5-Gram Sentences |
|---|---|
| 1 | episodes of your favorite haws |
| 2 | episodes of your favorite saws |
| 3 | episodes of your favorite hawk |
| 4 | episodes of your favorite shows |
| 5 | episodes of your favorite shawls |
| 6 | episodes of your favorite shays |
| 7 | episodes of your favorite shank |
| 8 | episodes of your favorite maws |

The pseudo-code of the error correction algorithm is given below.

```
FunctionErrorCorrection (candidates)
{
for(i←0 to i < N)  // process all candidate corrections
    {
    // concatenate together the ith candidate with the four
    preceding words
    // A is a global array containing the original ASR output text
L ←Concatenate(A[j-4] , A[j-3] , A[j-2] , A[j-1] ,
candidates[i] )

    // find L in Microsoft N-gram dataset and returns its frequency
    frequency[i] ← Search(MicrosoftDataset , L)

    i ← i+1
}

p←MaxFrequency(frequency)
// return the index p of the candidate whose L has the
highest frequency

    // return the correction for the ASR error
RETURN candidates[p]
}
```

# 7 Experiments & Results

For evaluation purposes, five different English articles each composed of around 100 words were read by five different speakers. Those articlesare from various domains including information technology, engineering, medicine, business, and sports. In sum, they consist of around500 words comprisingregular dictionary words, proper names, domain specific terms, special terminologies, and technical jargons. The ASR software used to perform the speech recognition is based on Microsoft Speech Application Programming Interface (SAPI 5.0) engine [24]. SAPI 5.0 is a freely-redistributable API developed by Microsoft and released in 2000 to allow the use of speech recognition and speech synthesis within Windows applications. Such applications include Microsoft Office, Microsoft Narrator, and Microsoft Speech Server.

The outcome of performing speech recognition using SAPIwasaround 106 total errors out of 500total words for the five articles, making the average of ASR errors around 21 errors per 100-words article. As a result, the error rate is 21% distributed as 14%non-word errors and 86% real-word errors.Table 4delineates the results obtained for SAPI including the total number of non-word and real-word errors.

TABLE 4
Results for the Microsoft SAPI

| Total Articles | Total Words | Total Errors | Average Total Errors | Total Non-Word Errors | Total Real-Word Errors |
|---|---|---|---|---|---|
| 5 | 500 | 106 | 106/5=21.2 | 15 | 91 |
| | | 21% | 21% | 14% of 106 | 86% of 106 |

Post-editing the obtained results using the proposed method resulted in 94 total errors being corrected successfully, among which 12 were non-word errors and 82 were real-word errors. As a result, around 89% of the total errors were corrected; around 80% of total non-word errors were corrected; and around 90% of total real-word errors were corrected successfully. Table 5 outlines the obtained test results for the proposed method.

TABLE 5
Results for the Proposed Method

| Total Errors 106 21% of 500 total words | | Non-Word Errors 15 14% of 106 | | Real-Word Errors 91 86% of 106 | |
|---|---|---|---|---|---|
| Corr-ected | Not Corrected | Corr-ected | Not Corrected | Corr-ected | Not Corrected |
| 94 | 12 | 12 | 3 | 82 | 9 |
| 89% of 106 | 11% of 106 | 80% of 15 | 20% of 15 | 90% of 91 | 10% of 91 |

In retrospect, the proposed method clearlyoutmatched the Microsoft SAPI engine as its error rate was around 21% for 500-words articles (106 errors out of 500 total words); while the error rate for the proposed method was around2.4% (12 errors out of 106 were left uncorrected, making the error rate 500*2.4%=12 errors out of 500 total words). These exceptional results are chiefly due to the bigamount of 5-gram data in Microsoft Web N-Gram dataset which were used by the proposed method as a dictionary to perform spelling detection and correction. As the content ofMicrosoft Web N-Gram dataset is minedfrom the World Wide Web, it is comprehensivelypacked with real-world data encompassing regular dictionary words, in addition to proper names, domain specific terms, special terminologies, uncommonacronyms, and technical jargons and expressions that can cover millions of words along with their possible sequences in the language.

## 8 CONCLUSIONS AND FUTURE WORK

This paper presented an original context-sensitive error correction method for detecting and correcting speech recognition non-word and real-word errors. The proposed system is based on Microsoft Web N-Gram dataset which incorporateslarge amount of real-world word sequences and n-gram statistics initially extracted from the World Wide Web, and necessary to effectively correct misspellings in any sort of text. Experiments conducted on several spoken articles showed a notable decrease in the ASR error rate. Practically, the error rate using the proposed method was around 2.4%, generating only 12 errors out of 500 total words; whereas, the error rate for other existing methods such as the SAPI engine was around 21%, generating 106 errors out of 500 total words.IntegratingMicrosoft Web N-Gram dataset into the proposed algorithms increased drastically the rate of error correction as this datasetholdsan extensivenumber of words and accurate statistics about word associations that almost cover the entire vocabulary of the language including regular words,proper names, domain specific terms, technical terminologies, scientific acronyms, special expressions, and a lot of word sequences that haveoriginallyoccurred on the web.

As for future work, the proposed methodis to be parallelized in an attempt to boostitsexecution time. Often, parallel algorithms, more specifically multithreaded algorithms can take the most out of multiprocessor large-scale computing machines to deliver very fast real-time computations, necessary to perform extensive error correction for long textat high speed and quality.

## ACKNOWLEDGMENT

This research was funded by the Lebanese Association for Computational Sciences (LACSC), Beirut, Lebanon under the "Web-Scale Speech Recognition Research Project – WSSRRP2011".